\documentclass{article}

\usepackage{microtype}
\usepackage{graphicx}
\usepackage{subfigure}
\usepackage{booktabs} 

\usepackage{graphicx}

\usepackage{hyperref}

\usepackage[accepted]{synsml2023}

\usepackage{amsmath}
\usepackage{amssymb}
\usepackage{mathtools}
\usepackage{amsthm}

\usepackage[capitalize,noabbrev]{cleveref}

\theoremstyle{plain}

\theoremstyle{definition}

\theoremstyle{remark}

\usepackage[textsize=tiny]{todonotes}

\usepackage{caption}
\synsmltitlerunning{OL-Transformer: A Fast and Universal Surrogate Simulator for Optical Multilayer Thin Film Structures}

\begin{document}

\twocolumn[
\synsmltitle{OL-Transformer: A Fast and Universal Surrogate Simulator for Optical Multilayer Thin Film Structures}



\synsmlsetsymbol{equal}{*}

\begin{synsmlauthorlist}
\synsmlauthor{Taigao Ma}{sch}
\synsmlauthor{Haozhu Wang}{comp}
\synsmlauthor{L. Jay Guo}{sch}
\end{synsmlauthorlist}

\synsmlaffiliation{comp}{Amazon Web Services, Illinois, USA}
\synsmlaffiliation{sch}{University of Michigan, Michigan, USA}

\synsmlcorrespondingauthor{L. Jay Guo}{guo@umich.edu}

\synsmlkeywords{Machine Learning}

\vskip 0.3in
]



\printAffiliationsAndNotice{}  

\begin{abstract}
Deep learning-based methods have recently been established as fast and accurate surrogate simulators for optical multilayer thin film structures. However, existing methods only work for limited types of structures with different material arrangements, preventing their applications towards diverse and universal structures. Here, we propose the \textbf{Opto-Layer (OL) Transformer} to act as a universal surrogate simulator for enormous types of structures. Combined with the technique of \textit{structure serialization}, our model can predict accurate reflection and transmission spectra for up to $10^{25}$ different multilayer structures, while still achieving a six-fold degradation in simulation time compared to physical solvers. Further investigation reveals that the general learning ability comes from the fact that our model first learns the physical embeddings and then uses the self-attention mechanism to capture the hidden relationship of light-matter interaction between each layer. 
\end{abstract}

\section{Introduction}
\label{introduction}
Optical multilayer thin film structure (shorten as “multilayer structure”) is a type of photonic structure that consists of multiple layers of different materials stacking on top of each other, with thickness typically ranging from tens of nanometers to a few micrometers. Because of the ease of fabrication, multilayer structures have been widely used in both scientific and industrial applications, including structural color \cite{wang2023structural}, photovoltaic \cite{liu2013efficient}, display devices \cite{zheludev2007life}, etc. To enable these applications, researchers need to first understand the physical relationship between a multilayer structure and corresponding optical properties, e.g., transmission and reflection. Traditional simulation methods including Transfer Matrix Methods (TMM) \cite{byrnes2016multilayer} and Rigorous Coupled Wave Analysis (RCWA) \cite{hugonin2021reticolo} use matrix algebra to analytically or semi-analytically calculate the reflection and transmission coefficients. However, these physics-based simulation methods are usually time-consuming. In addition, a new simulation needs to be performed from scratch when facing a different structure. Thus, the development of a fast simulation method becomes fundamental for multilayer structures applications.

Recently, researchers have started to use deep learning to accelerate the simulation process by leveraging the strong generalization ability, including Multi-Layer Perceptron (MLP) \cite{liu2018training}, MLP-Mixer \cite{deng2021benchmarking}, and transformer \cite{chen2023broadband}. Although obtaining the training dataset through physical simulation can take some time, such an investment is a one-time payment. Once trained, these neural networks are able to capture the general mapping from the space of structures to the space of optical properties, serving as a fast and computationally efficient surrogate model to replace physical simulations.

However, many existing surrogate models can only expedite the simulation of structures with fixed material arrangements, e.g., the three-layer structure of Ag/SiO$_{2}$/Ag in \cite{deng2021benchmarking} and the six-layer structure of MgF$_{2}$/SiO$_{2}$/Al$_{2}$O$_{2}$/TiO$_{2}$/Si/Ge in \cite{chen2023broadband}. This is because materials have dispersion, making them only accessible through categorical representations, instead of continuous variables. A universal method that can predict and simulate optical properties for diverse structures with different numbers of layers and varied materials arrangements is in great need. In this paper, we propose the Opto-Layer Transformer (OL-Transformer) as a universal surrogate model by leveraging the strong learning and generalization abilities of transformer. After training on a large dataset, our model can work as a fast and authentic surrogate solver for multilayer structures with up to 20 layers and 18 different materials, corresponding to a total of $18^{20}\sim 10^{25}$ different structures (this can be further expanded as our model is highly scalable). In addition, compared to physical simulators, our model also achieves a six-fold degradation in simulation time and can go to a $\sim$3800-fold speedup when using batch calculation.

\section{Related Works}
\label{Related Works}
Deep learning-based surrogate models have been used in many scientific applications to speed up the simulation and prediction, including molecular properties prediction \cite{broberg2022pre}, predicting the dynamics of physical phenomenon \cite{geneva2022transformers}, and meteorological predictions \cite{khorrami2021adapting}. They are also widely used for solving inverse problems, which seek to recover the causes given the observed results, including nanophotonic inverse design \cite{jiang2021deep}, chemical material inverse design \cite{fu2023material}, etc. Our work aims to solve the simulation of optical multilayer structures, with the goal of speeding up the prediction of optical properties.

There has been some work that leverages learned knowledge to deal with different structures. For example, \cite{qu2019migrating} used transfer learning to assist in the simulation of a 10-layer structure after learning on the 8-layer structure. \cite{kaya2019using} used transfer learning to help the optimization of a multi-layer solar cell. Meta-learning has also been applied to generalize the learning on different applications, including detector simulations and design \cite{zhang2020meta}, PINN-based 1D arc simulation \cite{zhong2023accelerating}, hydrogen storage materials simulations and design \cite{sun2021fingerprinting}, etc. However, these methods usually require adaptations to new datasets, restricting their applications to general purposes. We seek to find a universal surrogate model that works for as many different structures as possible through a single training.

\section{Methods}
\label{Methods}

\subsection{Problem Set}

\begin{figure}
\centering
\includegraphics[width=0.5\textwidth]{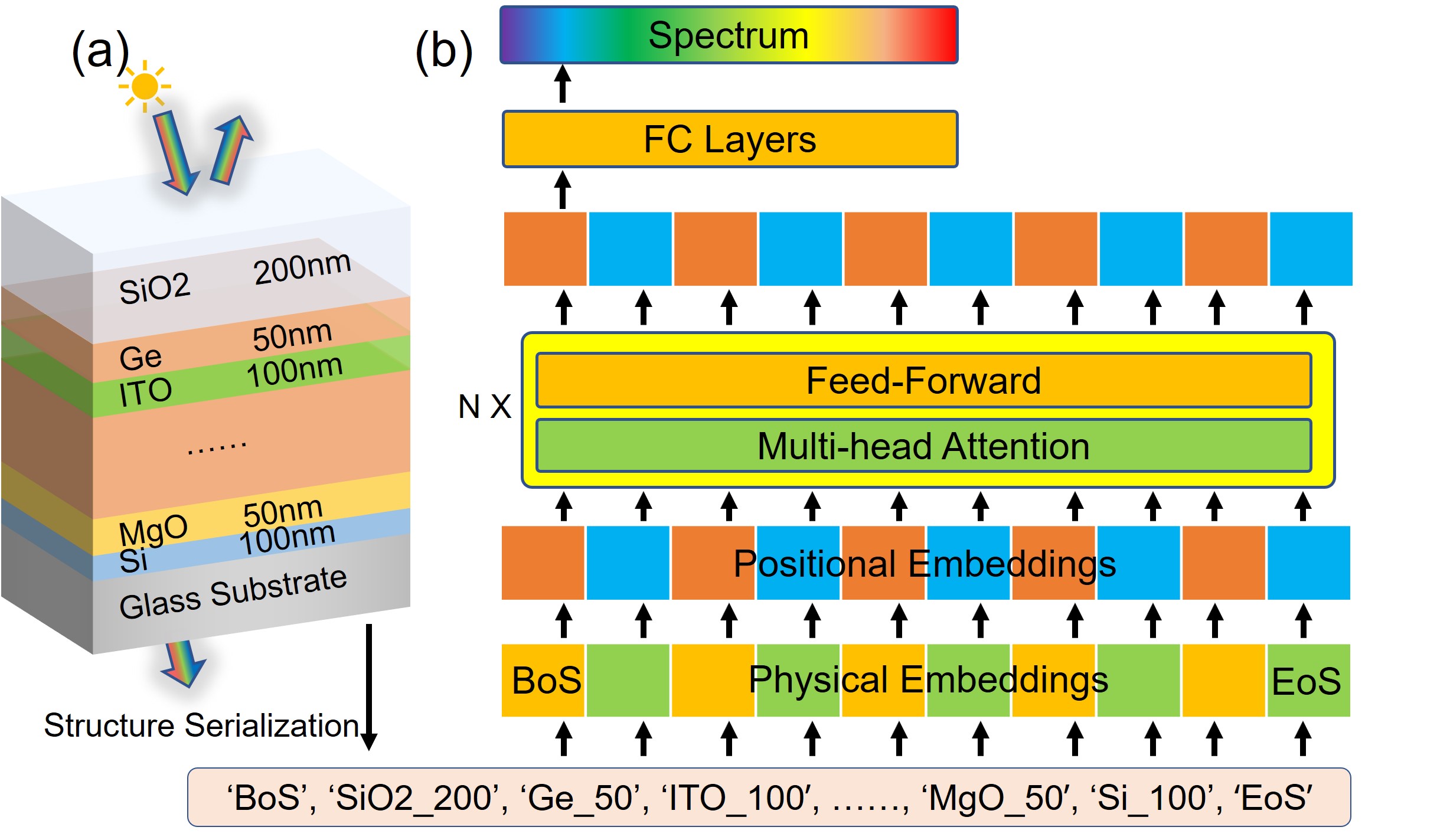}
\caption{\label{fig:model}(a) An example of a multilayer structure as well as its structure serialization. (b) The architecture of OL-Transformer. FC Layers: Fully Connected layers. }
\end{figure}

For a given multilayer structure with $N$ layers (see Fig. \ref{fig:model}a), we denote the material arrangements as $\mathbf{m}=\{m_1, m_2, \dots, m_N\}$ and the thickness sequence as $\mathbf{t}=\{t_1, t_2, \dots, t_N\}$. Here, $m_i, t_i$ refers to the material and thickness at the $i_{th}$ layer, respectively. $m_i\in\textbf{M}$ is a discrete variable that can take several distinct values from the material database $\textbf{M}$. Then, a multilayer structure can be described as $X=\{\textbf{m}, \textbf{t}\}\in\mathcal{X}$. A physical simulator $\mathcal{S}:\mathcal{X}\xrightarrow{}\mathcal{Y}$ maps the multilayer structure $X$ to the $d$-dimensional optical properties $Y=\{y_1, y_2, \dots, y_d\}\in\mathcal{Y}$ and works as an oracle for any type of material arrangements. Existing surrogate models $\hat{\textbf{S}}(\textbf{t}|\theta_{\textbf{m}}):\mathcal{X}\xrightarrow{}\hat{\mathcal{Y}}$ with learnable parameters $\theta_{\textbf{m}}$ take in the structure $X$ with different thickness $\mathbf{t}$ but with fixed material arrangement $\textbf{m}$ and output predicted optical properties $\hat{Y}=\{\hat{y_1}, \hat{y_2}, \dots, \hat{y_d}\}$. In this work, our surrogate model $\hat{S}(\textbf{m}, \textbf{t}|\theta):\mathcal{X}\xrightarrow{}\hat{\mathcal{Y}}$ with parameters $\theta$ wants to predict the optical properties for universal structures with different $\textbf{m}$ and different thickness $\mathbf{t}$. In this work, we consider predicting the transmission and reflection spectra from 400 nm to 1100 nm. Both spectra are discretized by 10 nm, making $d$=2$\times$71=142. Other types of optical properties can also be predicted similarly.

\subsection{Structure Serialization}
In order to expand the model’s capability towards versatile $\textbf{m}$, we combine with the recently developed technique called structure serialization \cite{ma2023optogpt}, where we use structure tokens to represent the material and thickness information $(m_i, t_i)$ at each layer simultaneously, similar to how Natural Language Processing (NLP) researchers tokenize language sentences. By appending multiple tokens one-by-one, we can convert a multilayer structure into a sequence of tokens that the transformer model can deal with. We also use a special token of ‘EoS’ (end of sequence) to enable the learning of structures with different numbers of layers (we set the maximum to be 20 layers). There are 18 different types of materials in our material database $\textbf{M}$, all of which are widely accessible in many nanofabrication centers. In addition, considering it is impossible to use an infinite number of tokens to describe a continuous thickness, we discretize the thickness by 10 nm and form 50 different choices from 10 nm to 500 nm. Therefore, there are 18$\times$50+1=901 tokens in our vocabulary and the total number of structures with different material arrangements expands to $18^{20}\sim10^{25}$. This method is scalable and can be used to include other materials and extend to more layers.

\subsection{Model Architecture}
Our model architecture is shown in Fig. \ref{fig:model}b, which is a standard encoder-only transformer that takes in the sequence of tokens. Each token will go through a physical embedding and positional embedding before being passed to the transformer model. The physical embedding plays a similar role as the input embedding in \cite{vaswani2017attention}. Similar to the class token in BERT \cite{devlin2018bert} and ViT \cite{dosovitskiy2020image}, we add a ‘BoS’ token (Beginning of Sequence) at the 0-th input position and treat its corresponding output from the encoder as the hidden representation for the input sentence. A fully connected layer is used to decode this output into the predicted spectra $\hat{Y}$. Our model is trained by minimizing the mean square error (MSE): $L=||Y-\hat{Y}||^2$. Masked language modeling is not used because each layer in the multilayer structure is equally important when predicting the optical properties. The training takes about one week on a single NVIDIA 3090 GPU. Table \ref{Parameter} lists important parameters of our model. 

\subsection{Dataset Generation}

Each training data contains a pair of multilayer structure and its spectra. During dataset generation, the multilayer structures are created by first randomly sampling the number of layers with an increasing ratio, and then uniformly sampling the material arrangements $\textbf{m}$ and the thickness sequence $\textbf{t}$. We then obtain their reflection and transmission spectra using TMM simulation. In total, we generate 10 M pairs of structures and spectra as the training dataset and another 1 M pairs for validation, which takes $\sim$1200 h for a single CPU. In the future analysis and comparison, we refer TMM as the accurate physical simulator and our OL-Transformer as the surrogate model.

\begin{table}[t]
\caption{Important parameters of our OL-Transformer}
\label{Parameter}
\vskip 0in
\begin{center}
\begin{small}
\begin{sc}
\begin{tabular}{cc}
    \toprule
    Name & Parameter \\
    \midrule
    Number of Encoder Block (N)  & 12 \\
    Number of Attention Head (A) & 16 \\
    Dim of Hidden States (H) & 1024 \\
    Number of Tokens    & 901 \\
    FC Layers           & 1024-1024-142 \\
    Number of Parameters    & 65 M\\
    \bottomrule
\end{tabular}
\end{sc}
\end{small}
\end{center}
\vskip -0.2in
\end{table}

\begin{table}[t]
\caption{Performance of our OL-Transformer. Batch size = 1000 for batch simulation.  MSE: Mean Square Error.}
\label{Performance}
\begin{center}
\begin{small}
\begin{sc}
\begin{tabular}{cccc}
    \toprule
    Attribute & TMM & Ours & Speedup \\
    \midrule
    Single Simulation (s) & 0.057 & 0.010 & $\sim$5.7 \\
    Batch Simulation (s) & - & 0.000015 & $\sim$3800 \\
    MSE & - & 0.000057 & - \\
    \bottomrule
\end{tabular}
\end{sc}
\end{small}
\end{center}
\vskip -0.2in
\end{table}

\section{Experiments}

\subsection{Simulaion Speedup}
In Table \ref{Performance}, we report the performance of simulation acceleration on the validation dataset. The physical simulation of TMM is evaluated on a single 2.4GHz CPU since there is no package available on GPU. Our model of OL-Transformer is evaluated on a single NVIDIA 3090. When making predictions for a single structure at one time, on average, our surrogate model can finish each simulation $\sim$5.7 times faster than the TMM. After using the GPU batch calculation (batch size = 1000), our model shows $\sim$3800 fold time improvement compared to TMM. Therefore, our model can be used as a faster simulator for multilayer thin film structures. This can be very helpful when a large number of simulations are needed, e.g., to inverse design or to understand the physical structure-property behaviors. 

\subsection{Universal Surrogate Simulator}
\begin{figure}[h]
\centering
\includegraphics[width=0.5\textwidth]{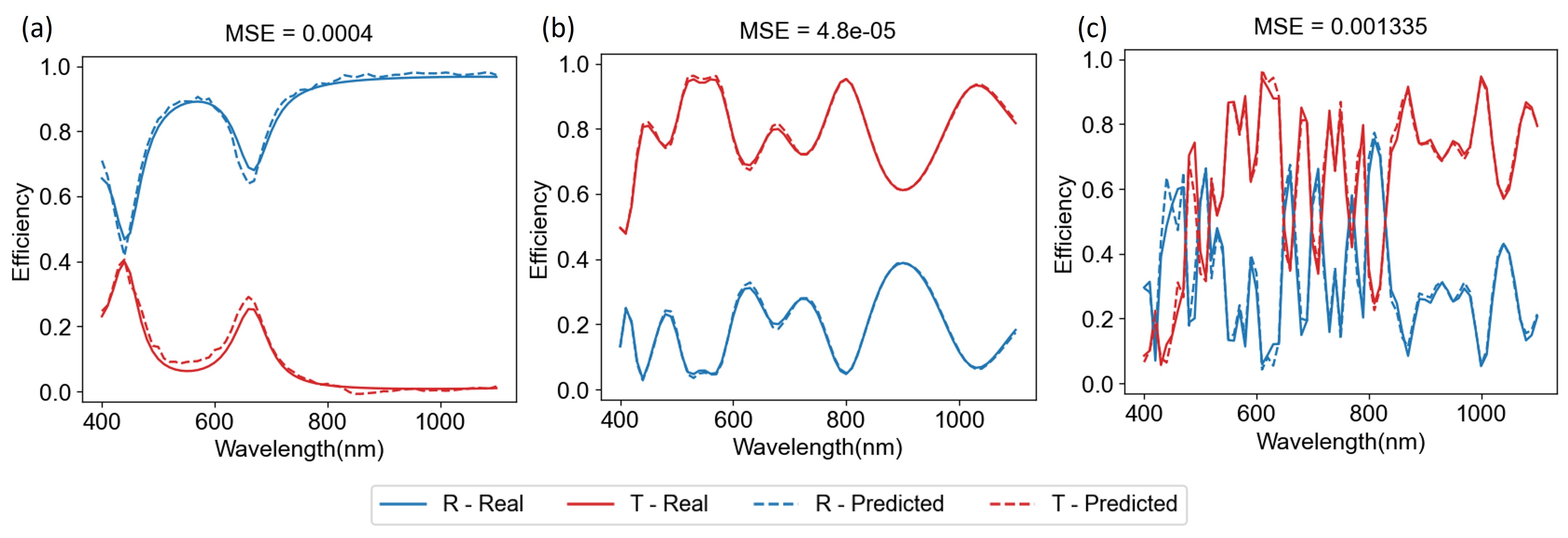}
\caption{\label{fig:examples} Three examples of predicting the transmission spectra (T) and reflection spectra (R). Title shows their MSE. }
\end{figure}

In principle, our method can extend up to $10^{25}$ different structures and work as a universal surrogate simulator. However, we cannot iterate and evaluate every one of them because there are so many. Therefore, in Table \ref{Performance}, we report the universal prediction ability on the validation dataset and evaluate the averaged MSE on $10^6$ different structures (since $10^6\ll 10^{25}$, each structure is a distinct type with different material arrangements). For a better comparison, we randomly select six structures with different $\textbf{m}$ and report their detailed prediction performance in Table \ref{Examples}. For each structure, we randomly generate 1000 structures with different $\textbf{t}$ and report their averaged MSE. Compared to the reported work which can only predict a specific structure, ours are versatile to different structures. Notice their MSE is listed only for reference because we are predicting different optical properties and cannot be compared directly. In Fig. \ref{fig:examples}, we also give three examples to visualize the difference between target spectra from TMM (real lines) and predicted spectra by our model (dashed lines). Based on the low MSE and close-to-simulator spectra, we demonstrate that our model exhibits a strong generalization ability to predict the spectra of universal types of structures, significantly expanding the capabilities of existing surrogate models.  

\begin{table*}[t]
\caption{Prediction performance on six different structures. Layers: The total number of layers in the given type of multilayer structure. Reported MSE: The reported MSE in existing work is summarized only for reference. We cannot directly compare them because the predicted optical properties are different.}
\label{Examples}
\vskip -1in
\begin{center}
\begin{small}
\begin{sc}
\begin{tabular}{cccc}
    \toprule
    Description of Multilayer Structure & Layers & MSE & Reported MSE\\
    \midrule
    $\mathrm{Ag/SiO_2/Ag}$ & 3 & $8.2\times10^{-4}$ & $\sim10^{-5}$\cite{deng2021benchmarking}\\
    $\mathrm{MgF_2/SiO_2/Al_2O_3/TiO_2/Si/Ge}$ & 6 & $4.9\times10^{-4} $ & $\sim10^{-6}$\cite{chen2023broadband}  \\
    $\mathrm{SiO_2/Si_3N_4/SiO_2/Si_3N_4/SiO_2/Si_3N_4}$ & 6 & $2.6\times10^{-5} $ & - \\
    $\mathrm{TiO_2/SiO_2/Al_2O_3/Si_3N_4/ZnO/ZnS/ITO/HfO_2/Si}$ & 9 & $2.5\times10^{-5} $ & - \\
    $\mathrm{ZnS/TiO_2/MgO/ZnS/Si_3N_4/ITO/SiO_2/TiO_2/Ta_2O_5/ZnO/Al_2O_3/Ag}$ & 12 & $1.3\times10^{-5} $ & - \\
    $
    
     \begin{aligned}[c] 
     \mathrm{SiO_2/Si_3N_4/SiO_2/Si_3N_4/SiO_2/Si_3N_4/SiO_2/Si_3N_4/SiO_2/Si_3N_4/} \\ \mathrm{SiO_2/Si_3N_4/SiO_2/Si_3N_4/SiO_2/Si_3N_4/SiO_2/Si_3N_4/SiO_2/Si_3N_4}

         \end{aligned}
         
         $ & 20 & $1.1\times10^{-3} $ & $\sim10^{-4}$\cite{liu2018training} \\
    \bottomrule
\end{tabular}
\end{sc}
\end{small}
\end{center}
\vskip -0.2in
\end{table*}

\subsection{Understanding the Universal Learning Ability}
\begin{figure}[h]
\centering
\includegraphics[width=0.5\textwidth]{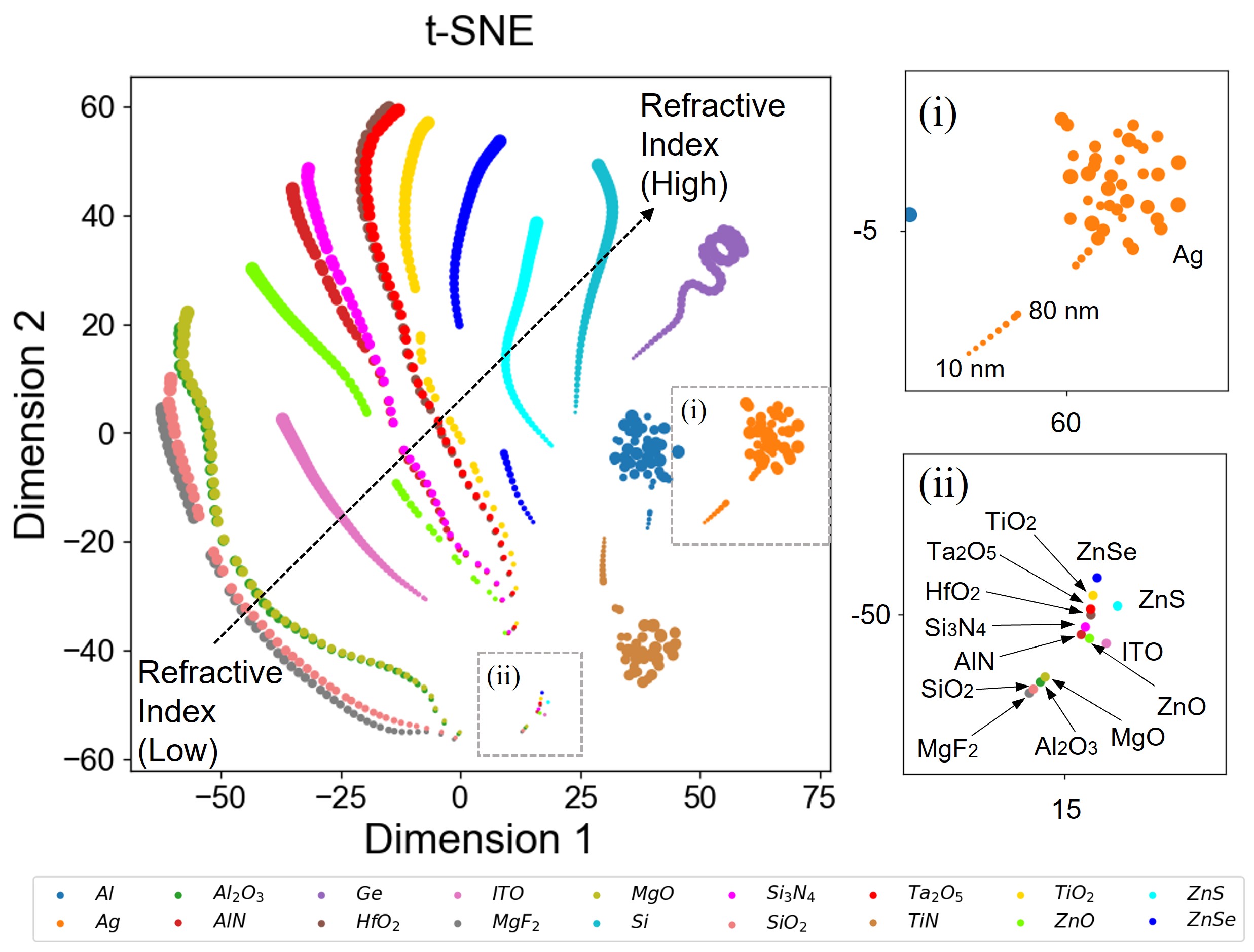}
\caption{\label{fig:tsne} The 2D visualization of the hidden space using t-SNE to reduce dimension.}
\end{figure}

To understand why our model exhibits a strong generalized learning ability, we first use t-SNE to reduce the high-dimensional physical embeddings for each token into two dimensions and visualize the results in Fig. \ref{fig:tsne}. These tokens referring to the same material are marked as the same color. Clearly, we can see a transition of materials from the low refractive index region (lower left) to the high refractive index region (upper right). For each material, the increasing size of dots corresponds to the increasing thickness, e.g., 10 nm for the smallest dot and 500 nm for the largest dot. Because our model only takes in the tokens, instead of the thickness or material's properties, these observations demonstrate that our model exhibits the ability to recognize and recover the material and thickness information using hidden representations. Apart from this, intrinsic physics can also be obtained from Fig. \ref{fig:tsne}. For example, these dots representing metals, e.g, Al, Ag (see zoomed view (i)), are distinguishable at small thickness but cluster together at greater thickness. This makes sense as metals with thickness greater than the penetration depth have no impact on light-matter interaction. In addition, most dielectric materials with small thickness are clustered together (see zoomed view (ii)), which reminds us of the fact that thin dielectric materials are hard to distinguish because they all have a similar impact on light propagation.  

\begin{figure}[h]
\centering
\includegraphics[width=0.5\textwidth]{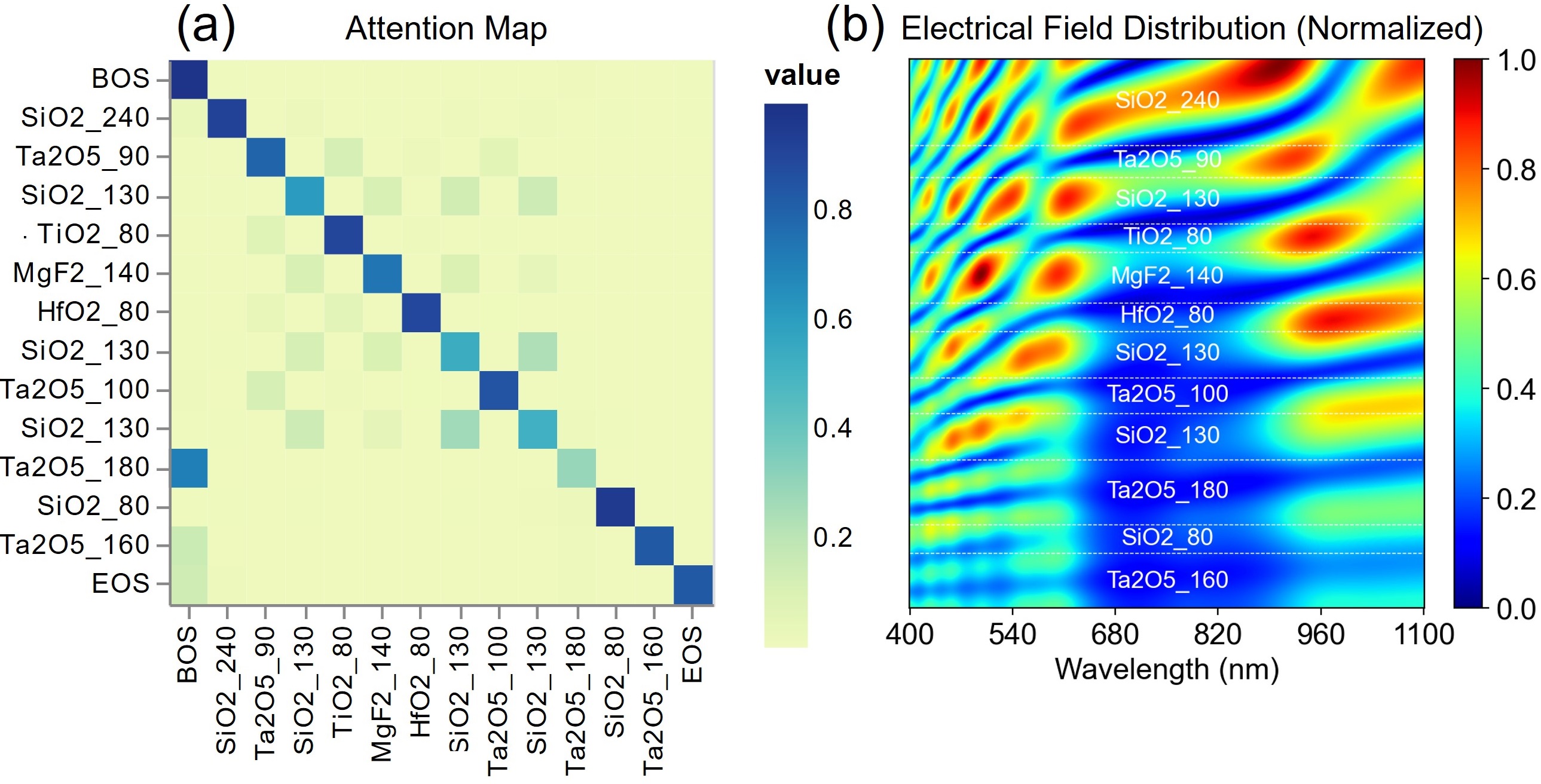}
\caption{\label{fig:attention} (a) Visualization of attention map (Block 0, Head 15). (b) The electrical field distribution. }
\end{figure}

To further understand how our model leverages these learned embeddings for surrogate simulation, we then visualize one of the attention maps for a given multilayer structure in Fig. \ref{fig:attention}a. We also simulate and visualize the electrical field distribution from 400 nm to 1100 nm by solving the multilayer system through TMM, as shown in Fig. \ref{fig:attention}b. The attention map in Fig. \ref{fig:attention}a exhibits an obvious alternating pattern on off-diagonal elements, suggesting a high correlation in these nearby layers. Interestingly, such alternating patterns can also be observed in Fig. \ref{fig:attention}b, where the electrical field illustrates how light interacts and propagates among each layer. Intuitively, self-attention can be treated as the analogy of physical interactions inside the transformer model.  

In summary, the empirical understanding of the strong universal learning ability can be explained by 1) our model learns the unique embeddings with intrinsic physical meanings for all structure tokens and 2) our model learns to use self-attention to represent the light-matter interaction among each layer. Although there are up to $10^{25}$ different material arrangements, all of them can be reconstructed by 901 structure tokens. Our model learns to decompose the prediction of each type of structure into the two-step learning of physical embeddings and self-attention.

\section{Conclusion}
In this paper, we introduce \textbf{OL-Transformer}, a fast and universal surrogate solver for simulating the transmission and reflection spectra in optical multilayer thin film structures. Compared to existing surrogate models, our model can extend the learning capabilities from limited structures to $10^{25}$ different structures without adaptation used in transfer learning or meta-learning. In addition, our model still achieves a six-fold speedup compared to numerical simulation, with a potential 3800-fold when using batch calculations, facilitating the downstream applications including inverse design and understanding the structure-property behaviors. Our work expands the existing transformer applications from mainstream NLP and CV to optical simulations and demonstrates that transformer architecture (embedding and self-attention) is also an effective learner for optical physics.

\section{Potential Broader Impact}

In optics and photonics, understanding the structure-property relationship is vital for developing optical devices and photonic applications. By leveraging the strong generalization ability of transformer, our model can serve as a fast and universal surrogate simulator for predicting the optical properties of optical multilayer thin film structures, e.g., transmission and reflection spectra, providing an easy and straightforward to explore the light-matter interaction and inverse design for specific structures to satisfy desired optical properties. Our method can be easily scaled to predict other types of optical properties, including angled-resolved spectra and structure color, and include more materials with more complex layered structures. In addition, the way we serialize the multilayer structure as a sequence of tokens can also be directly applied to other types of photonic structures, including meta-grating structures and free-form metasurfaces, with the potential to speed up the simulation and prediction of these structures with higher complexity. On the other hand, our work also demonstrates that transformer has a strong capability of capturing intrinsic physical knowledge and using them for predicting physical behaviors, which can inspire future work to leverage transformer to solve complicated and non-trivial physical problems.

\nocite{langley00}

\bibliography{example_paper}

\begin{thebibliography}{23}
\providecommand{\natexlab}[1]{#1}
\providecommand{\url}[1]{\texttt{#1}}
\expandafter\ifx\csname urlstyle\endcsname\relax
  \providecommand{\doi}[1]{doi: #1}\else
  \providecommand{\doi}{doi: \begingroup \urlstyle{rm}\Url}\fi

\bibitem[Broberg et~al.(2022)Broberg, B{\aa}nkestad, and
  Ylip{\"a}{\"a}]{broberg2022pre}
Broberg, J., B{\aa}nkestad, M., and Ylip{\"a}{\"a}, E.
\newblock Pre-training transformers for molecular property prediction using
  reaction prediction.
\newblock \emph{arXiv preprint arXiv:2207.02724}, 2022.

\bibitem[Byrnes(2016)]{byrnes2016multilayer}
Byrnes, S.~J.
\newblock Multilayer optical calculations.
\newblock \emph{arXiv preprint arXiv:1603.02720}, 2016.

\bibitem[Chen et~al.(2023)Chen, Gao, Li, Yan, Ou, Ma, and
  Zhu]{chen2023broadband}
Chen, W., Gao, Y., Li, Y., Yan, Y., Ou, J.-Y., Ma, W., and Zhu, J.
\newblock Broadband solar metamaterial absorbers empowered by transformer-based
  deep learning.
\newblock \emph{Advanced Science}, pp.\  2206718, 2023.

\bibitem[Deng et~al.(2021)Deng, Dong, Ren, Khatib, Soltani, Tarokh, Padilla,
  and Malof]{deng2021benchmarking}
Deng, Y., Dong, J., Ren, S., Khatib, O., Soltani, M., Tarokh, V., Padilla, W.,
  and Malof, J.
\newblock Benchmarking data-driven surrogate simulators for artificial
  electromagnetic materials.
\newblock In \emph{Thirty-fifth Conference on Neural Information Processing
  Systems Datasets and Benchmarks Track (Round 2)}, 2021.

\bibitem[Devlin et~al.(2018)Devlin, Chang, Lee, and Toutanova]{devlin2018bert}
Devlin, J., Chang, M.-W., Lee, K., and Toutanova, K.
\newblock Bert: Pre-training of deep bidirectional transformers for language
  understanding.
\newblock \emph{arXiv preprint arXiv:1810.04805}, 2018.

\bibitem[Dosovitskiy et~al.(2020)Dosovitskiy, Beyer, Kolesnikov, Weissenborn,
  Zhai, Unterthiner, Dehghani, Minderer, Heigold, Gelly,
  et~al.]{dosovitskiy2020image}
Dosovitskiy, A., Beyer, L., Kolesnikov, A., Weissenborn, D., Zhai, X.,
  Unterthiner, T., Dehghani, M., Minderer, M., Heigold, G., Gelly, S., et~al.
\newblock An image is worth 16x16 words: Transformers for image recognition at
  scale.
\newblock \emph{arXiv preprint arXiv:2010.11929}, 2020.

\bibitem[Fu et~al.(2023)Fu, Wei, Song, Li, Xin, Omee, Dong, Siriwardane, and
  Hu]{fu2023material}
Fu, N., Wei, L., Song, Y., Li, Q., Xin, R., Omee, S.~S., Dong, R., Siriwardane,
  E. M.~D., and Hu, J.
\newblock Material transformers: deep learning language models for generative
  materials design.
\newblock \emph{Machine Learning: Science and Technology}, 4\penalty0
  (1):\penalty0 015001, 2023.

\bibitem[Geneva \& Zabaras(2022)Geneva and Zabaras]{geneva2022transformers}
Geneva, N. and Zabaras, N.
\newblock Transformers for modeling physical systems.
\newblock \emph{Neural Networks}, 146:\penalty0 272--289, 2022.

\bibitem[Hugonin \& Lalanne(2021)Hugonin and Lalanne]{hugonin2021reticolo}
Hugonin, J.~P. and Lalanne, P.
\newblock Reticolo software for grating analysis.
\newblock \emph{arXiv preprint arXiv:2101.00901}, 2021.

\bibitem[Jiang et~al.(2021)Jiang, Chen, and Fan]{jiang2021deep}
Jiang, J., Chen, M., and Fan, J.~A.
\newblock Deep neural networks for the evaluation and design of photonic
  devices.
\newblock \emph{Nature Reviews Materials}, 6\penalty0 (8):\penalty0 679--700,
  2021.

\bibitem[Kaya \& Hajimirza(2019)Kaya and Hajimirza]{kaya2019using}
Kaya, M. and Hajimirza, S.
\newblock Using a novel transfer learning method for designing thin film solar
  cells with enhanced quantum efficiencies.
\newblock \emph{Scientific reports}, 9\penalty0 (1):\penalty0 1--10, 2019.

\bibitem[Khorrami et~al.(2021)Khorrami, Simek, Cheung, Veillette, Dangovski,
  Rugina, Soljacic, and Agrawal]{khorrami2021adapting}
Khorrami, P., Simek, O., Cheung, B., Veillette, M., Dangovski, R., Rugina, I.,
  Soljacic, M., and Agrawal, P.
\newblock Adapting deep learning models to new meteorological contexts using
  transfer learning.
\newblock In \emph{2021 IEEE International Conference on Big Data (Big Data)},
  pp.\  4169--4177. IEEE, 2021.

\bibitem[Langley(2000)]{langley00}
Langley, P.
\newblock Crafting papers on machine learning.
\newblock In Langley, P. (ed.), \emph{Proceedings of the 17th International
  Conference on Machine Learning (ICML 2000)}, pp.\  1207--1216, Stanford, CA,
  2000. Morgan Kaufmann.

\bibitem[Liu et~al.(2018)Liu, Tan, Khoram, and Yu]{liu2018training}
Liu, D., Tan, Y., Khoram, E., and Yu, Z.
\newblock Training deep neural networks for the inverse design of nanophotonic
  structures.
\newblock \emph{Acs Photonics}, 5\penalty0 (4):\penalty0 1365--1369, 2018.

\bibitem[Liu et~al.(2013)Liu, Johnston, and Snaith]{liu2013efficient}
Liu, M., Johnston, M.~B., and Snaith, H.~J.
\newblock Efficient planar heterojunction perovskite solar cells by vapour
  deposition.
\newblock \emph{Nature}, 501\penalty0 (7467):\penalty0 395--398, 2013.

\bibitem[Ma et~al.(2023)Ma, Wang, and Guo]{ma2023optogpt}
Ma, T., Wang, H., and Guo, L.~J.
\newblock Optogpt: A foundation model for inverse design in optical multilayer
  thin film structures.
\newblock \emph{arXiv preprint arXiv:2304.10294}, 2023.

\bibitem[Qu et~al.(2019)Qu, Jing, Shen, Qiu, and Soljacic]{qu2019migrating}
Qu, Y., Jing, L., Shen, Y., Qiu, M., and Soljacic, M.
\newblock Migrating knowledge between physical scenarios based on artificial
  neural networks.
\newblock \emph{ACS Photonics}, 6\penalty0 (5):\penalty0 1168--1174, 2019.

\bibitem[Sun et~al.(2021)Sun, DeJaco, Li, Tang, Glante, Sholl, Colina, Snurr,
  Thommes, Hartmann, et~al.]{sun2021fingerprinting}
Sun, Y., DeJaco, R.~F., Li, Z., Tang, D., Glante, S., Sholl, D.~S., Colina,
  C.~M., Snurr, R.~Q., Thommes, M., Hartmann, M., et~al.
\newblock Fingerprinting diverse nanoporous materials for optimal hydrogen
  storage conditions using meta-learning.
\newblock \emph{Science Advances}, 7\penalty0 (30):\penalty0 eabg3983, 2021.

\bibitem[Vaswani et~al.(2017)Vaswani, Shazeer, Parmar, Uszkoreit, Jones, Gomez,
  Kaiser, and Polosukhin]{vaswani2017attention}
Vaswani, A., Shazeer, N., Parmar, N., Uszkoreit, J., Jones, L., Gomez, A.~N.,
  Kaiser, {\L}., and Polosukhin, I.
\newblock Attention is all you need.
\newblock \emph{Advances in neural information processing systems}, 30, 2017.

\bibitem[Wang et~al.(2023)Wang, Liu, Wang, Li, Guo, and
  Zhang]{wang2023structural}
Wang, D., Liu, Z., Wang, H., Li, M., Guo, L.~J., and Zhang, C.
\newblock Structural color generation: from layered thin films to optical
  metasurfaces.
\newblock \emph{Nanophotonics}, 12\penalty0 (6):\penalty0 1019--1081, 2023.

\bibitem[Zhang et~al.(2020)Zhang, He, Li, Wen, and Jin]{zhang2020meta}
Zhang, J., He, Y., Li, Y.-W., Wen, C.-K., and Jin, S.
\newblock Meta learning-based mimo detectors: Design, simulation, and
  experimental test.
\newblock \emph{IEEE Transactions on Wireless Communications}, 20\penalty0
  (2):\penalty0 1122--1137, 2020.

\bibitem[Zheludev(2007)]{zheludev2007life}
Zheludev, N.
\newblock The life and times of the led—a 100-year history.
\newblock \emph{Nature photonics}, 1\penalty0 (4):\penalty0 189--192, 2007.

\bibitem[Zhong et~al.(2023)Zhong, Wu, and Wang]{zhong2023accelerating}
Zhong, L., Wu, B., and Wang, Y.
\newblock Accelerating physics-informed neural network based 1d arc simulation
  by meta learning.
\newblock \emph{Journal of Physics D: Applied Physics}, 56\penalty0
  (7):\penalty0 074006, 2023.

\end{thebibliography}
\bibliographystyle{synsml2023}

\newpage
\appendix
\onecolumn


\end{document}